\documentclass{ieeeaccess}

\usepackage{textcomp}

\def\BibTeX{{\rm B\kern-.05em{\sc i\kern-.025em b}\kern-.08em
    T\kern-.1667em\lower.7ex\hbox{E}\kern-.125emX}}

\usepackage[noadjust]{cite}
\usepackage[pdftex]{graphicx}
\graphicspath{{../pdf/}{../jpeg/}}
\DeclareGraphicsExtensions{.pdf,.jpeg,.png}
\usepackage{calrsfs}
\usepackage{amsmath,amssymb,amsfonts}
\usepackage{hhline}
\usepackage{mathtools}
\usepackage{algpseudocode}
\usepackage{algorithm}
\usepackage{array}
\usepackage[caption=false,font=footnotesize]{subfig}
\usepackage{stfloats}
\usepackage{float}
\usepackage{url}
\usepackage{comment}
\usepackage{multirow}
\usepackage{mwe}
\usepackage{color}

\begin{document}
\history{Date of publication January 20, 2021, date of current version January 27, 2021.}
\doi{10.1109/ACCESS.2021.3052996}
\vol{9}

\bstctlcite{IEEEexample:BSTcontrol}
\title{Neural Architecture Search by Estimation of Network Structure Distributions}

\author{\uppercase{Anton Muravev}\authorrefmark{1}, \IEEEmembership{Member,~IEEE}, \uppercase{Jenni Raitoharju}\authorrefmark{1,2}, \IEEEmembership{Member,~IEEE} and \uppercase{Moncef Gabbouj} \authorrefmark{1}, \IEEEmembership{Fellow,~IEEE}}
\address[1]{Faculty of Information Technology and Communication Sciences, Tampere University, FI-33100 Tampere, Finland.}
\address[2]{Programme for Environmental Information, Finnish Environment Institute, FI-40500 Jyväskylä, Finland}
\tfootnote{This  work  was supported by the European Union’s Horizon 2020 research and innovation programme under grant agreement  No  871449  (OpenDR). }

\markboth
{Muravev \headeretal: Neural Architecture Search by Estimation of Network Structure Distributions}
{Muravev \headeretal: Neural Architecture Search by Estimation of Network Structure Distributions}

\corresp{Corresponding author: Anton Muravev (e-mail: anton.muravev@tuni.fi).}

\begin{abstract}
The influence of deep learning is continuously expanding across different domains, and its new applications are ubiquitous. The question of neural network design thus increases in importance, as traditional empirical approaches are reaching their limits. Manual design of network architectures from scratch relies heavily on trial and error, while using existing pretrained models can introduce redundancies or vulnerabilities. Automated neural architecture design is able to overcome these problems, but the most successful algorithms operate on significantly constrained design spaces, assuming the target network to consist of identical repeating blocks. While such approach allows for faster search, it does so at the cost of expressivity. We instead propose an alternative probabilistic representation of a whole neural network structure under the assumption of independence between layer types. Our matrix of probabilities is equivalent to the population of models, but allows for discovery of structural irregularities, while being simple to interpret and analyze. We construct an architecture search algorithm, inspired by the estimation of distribution algorithms, to take advantage of this representation. The probability matrix is tuned towards generating high-performance models by repeatedly sampling the architectures and evaluating the corresponding networks, while gradually increasing the model depth. Our algorithm is shown to discover non-regular models which cannot be expressed via blocks, but are competitive both in accuracy and computational cost, while not utilizing complex dataflows or advanced training techniques, as well as remaining conceptually simple and highly extensible.
\end{abstract}


\titlepgskip=-15pt

\maketitle

%
\IEEEpeerreviewmaketitle

\section{Introduction}
\IEEEPARstart{T}{he} recent successes of deep learning have attracted significant interest in numerous fields of knowledge \cite{LeCun2015}. Computer vision in particular has witnessed the development of multiple successful models, based on convolutional neural networks (CNNs), for tasks such as classification \cite{He2015, Huang2017a}, semantic segmentation \cite{Chen2016}, and detection \cite{NIPS2015_5638}. While the growth of deep learning solutions over the years is impressive, their adoption brings many significant challenges of both theoretical and practical nature. In addition to well-known problems such as overfitting or vanishing gradients, which have been subjects of extensive research over the years, new issues are being discovered, which are not yet fully understood. For example, the lack of interpretability of decisions made by deep models \cite{Lipton2016, Charles2018} is a difficult problem to tackle, but has attracted increasing research attention recently \cite{Montavon2018}. Further concerns have been raised regarding secure practical use of common deep models, as they were shown to be vulnerable to attacks utilizing malicious data \cite{Nguyen2015}.

One aspect of the neural networks, intricately tied to these challenges, is the architectural design: the choice of layer count, connection patterns, neuron operations, and their hyperparameters (convolution filter sizes, channel depth). It is well-known that some structural choices are associated with training difficulties; for example, a depth increase causes the vanishing gradient problem \cite{He2016}. Meanwhile, on a system level, the design guidelines of creating a deep network for a particular practical problem are not well established. The network design task thus becomes a time-demanding process, involving extensive trial and error. In practice, this issue is commonly avoided by using an already established pre-trained model of the same or related data domain as a feature extractor \cite{Weiss2016}. While effective, the latter approach presents problems of its own. Pre-trained models tend to be large and can lead to resource-consuming, largely redundant systems, whereas a too small network may produce insufficient accuracy for the problem at hand. Specific features of the data may require specific layer types to be fully exploited \cite{Muravev2018}. Pre-trained models can carry undesired biases from their original datasets \cite{Tommasi2017}. Additionally, sharing the foundation means that such systems will naturally be more vulnerable to adversarial attacks. A promising way to avoid these problems lies in developing appropriate methods for automated task-specific network design.

The idea of automated neural network design dates back to the early 1990s \cite{Whitley1990a}. The following decade saw a large volume of research on this problem, primarily focusing on evolutionary algorithms as solvers, both due to their gradient-free nature and shared biological inspirations \cite{XinYao1999}. This family of approaches would later be coined \emph{neuroevolution} \cite{Stanley2002a}. The research continued into the 2000s, with both improved evolutionary algorithms \cite{Stanley2002a, Stanley2009a} and other metaheuristic approaches, such as particle swarm optimization \cite{Kiranyaz2009}, as the search method. However, all of these algorithms share the need to perform many evaluations of intermediate solutions, which, in the case of automated architecture design, requires training numerous candidate networks from scratch. Therefore, these approaches were computationally restricted to rather limited model complexity, and their practical applications remained primarily in control tasks and robotics, where these limitations have a smaller impact \cite{Floreano2008}.

The advent of deep learning, where training a single model can take days or weeks, caused manual design to once again become the primary approach. However, as architectural discoveries paved the way to models with a small number of parameters and superior performance \cite{He2015, Huang2017a}, the interest in automated design reemerged, taking advantage of both evolutionary optimization \cite{Xie2017, Real2017, Zhang2018a, Real2018a, Sun2019a} and newer approaches, such as reinforcement learning \cite{Zoph2016, Zoph2017, Pham2018a}. The computational demand remains a major limitation and has hence been the focus of most recent works in the area \cite{Pham2018a, Liu2018, Elsken2019}. 

Another concern is the growing \emph{semantic} complexity of such algorithms. While they may yield successful architectures \cite{Real2018a}, their search behaviour is hard to analyze, which obscures the effects (whether positive or negative) of individual algorithmic steps and hinders comparisons. For instance, Zoph \textit{et~al.} \cite{Zoph2017} proposed a reduced search space: the network is represented as a repeating sequence of cells of a few types, where the internal structure of cells of the same type is identical and subject to optimization. Consequently, this design space has been adopted by a multitude of other approaches (see \cite{Pham2018a, Real2018a, Liu2018a}), as it allowed significant speedups. However, the random search has recently been shown to be highly competitive in this space as well, suggesting that previous successes may come not from algorithm specifics but from the expressivity constraints \cite{Sciuto2019}. 

Despite the variety of the proposed solutions for the architecture search problem, the majority of them rely on the repeating blocks, cells, or motifs. While effective in reducing the search space and thus (indirectly) the computation required, such an approach is by design biased towards deeper and uniformly structured models. This brings them closer to handcrafted networks, but limits the possibilities to discover irregular architectures with non-repetitive patterns, which can potentially be superior in performance or have other valuable properties, such as faster inference time or lower memory footprint.

We propose a conceptually simple and extensible architecture search method that runs on a less structured, dynamically growing search space. Our solution is based on the estimation of distribution algorithms. We specifically draw from Population-Based Incremental Learning \cite{Baluja1994} and Univariate Marginal Distribution Algorithm  \cite{Muhlenbein1997}. We utilize a set of discrete probability distributions to describe the choice of layers in a feedforward deep network, assuming their independence. Together they form a \emph{network prototype}, which is then iteratively updated by sampling and evaluating network models, until convergence is reached. The contributions of this paper can be summarized as follows:
\begin{itemize}
    \item We propose a probabilistic representation of deep neural networks by expressing their structure as a set of layer type probabilities. A single prototype of the proposed form corresponds to not a single network, but a family of models, which can span entire regions of the design space. This representation is less constrained compared to cells in that it does not force repeating patterns to appear, while still allowing the regularity to emerge by itself (if it proves competitive).
    \item We propose a CNN architecture search method based on the optimization of the above prototype, denoted Architecture Search by Estimation of network structure Distributions (ASED). As candidate networks are sampled and individual probabilities converge to their extreme values, the algorithm naturally transitions from global to local search, avoiding suboptimal areas. Unlike cell-based methods, we dynamically expand the search space to gradually explore deeper and deeper architectures over the runtime.
    \item We propose additional techniques to introduce non-linear connectivity patterns to solutions and to control the speed of search convergence. Evaluation results confirm that these techniques lead to performance improvements.
    \item We experimentally demonstrate the comparable performance of our method (in our growing search space) to existing approaches in terms of both model performance and computational requirements. We notably use only feedforward structures without explicitly defined repeated motifs or advanced techniques for regularization and data pre-processing.
\end{itemize}

The rest of the paper is structured as follows. In Section~2, we review the related developments in the field of neural architecture optimization as well as elaborate on our inspirations. Section~3 describes our approach to architecture search and the additional related techniques. Section~4 contains the experimental results and their analysis. Finally, Section~5 concludes the work and outlines some potential future studies.

\section{Related Work}

Ever since the wider adoption of multilayer perceptron structures \cite{Rumelhart1988}, their architectures became subject to optimization. While general neural network design involved heuristic rules and empirical tests, a promising alternative was found in evolutionary optimization methods due to their ability to solve problems defined only by the target function, without requiring any gradient information \cite{Back1997}\cite{Luke2013}. Early works on neuroevolution included \cite{Gomez1997, Yao1997, Gomez1999}. However, all of these works shared common issues of prohibitive computational requirements and lack of robustness due to the highly noisy nature of the search space \cite{XinYao1999}.

The small-scale neuroevolution reached a new peak when NEAT (NeuroEvolution of Augmented Topologies) \cite{Stanley2002a} was introduced in 2002. The techniques that made NEAT differ from its predecessors are historical gene markings, allowing for straightforward and meaningful crossover, and speciation with fitness sharing, which allows promising individuals to more consistently reach their full potential. Despite the advantages and the flexibility it offered, NEAT remained limited to small-scale applications, such as control tasks with limited inputs (a problem which would later be tackled within reinforcement learning). Multiple subsequent variants of NEAT \cite{Stanley2007a,Stanley2009a,Risi2012} aimed at efficient generation and representation of more complex networks with repeatable structural patterns. For example, HyperNEAT \cite{Stanley2009a} encodes network architectures via a metric space, called \emph{substrate}, where every neuron is mapped to a point with fixed coordinates. A separate hypermodel, called a Compositional Pattern-Producing Network (CPPN), takes as inputs the coordinates of a pair of neurons and outputs the connection weight between them, thus defining a network structure. An evolved substrate variant of HyperNEAT, or es-HyperNEAT \cite{Risi2012}, avoids the need to explicitly define node geometry, allowing for discovery of a wider scope of structures. However, despite the greater representational expression of these methods, the overarching problems remained and the use was limited to specific small-scale applications \cite{DAmbrosio2014}.

The resurgence of interest for architecture optimization started in 2016, after the introduction of reinforcement learning driven Neural Architecture Search (NAS) in \cite{Zoph2016}. An LSTM-based recurrent neural network (a controller) is trained to output sequences of tokens (symbols), which correspond to specific values of convolutional layers' parameters, such as filter size, count, and stride. The resulting neural network can be trained and evaluated. The controller can then be updated by the REINFORCE rule to maximize the expected accuracy of the generated networks. The major weakness of NAS is the computational cost of over 22000 GPU days on a standard CIFAR-10 image classification dataset. The follow-up work NAS-Net \cite{Zoph2017} represents the target network as a predefined sequence of repeating elements, known as \emph{cells}. Each cell type shares the same internal structure, which is optimized in a graph form and can contain different convolution and pooling operations. During the search process, the total number of cells in the network is reduced to speed up computation, while the final discovered architecture is evaluated in a full-length sequence. Such a reduction of the design space has proven effective in guiding the search, thus boosting the accuracy and reducing the running time to 2000 GPU days, and has since been used in other works, despite the limitations of the representation. Efficient NAS, or ENAS \cite{Pham2018a}, achieves further speed-up (to less than 16 GPU hours) at the cost of some accuracy loss. It utilizes \emph{weight sharing}, where the convolutional filter weights are identical between the cells and depend only on the position of the corresponding edge in the structural graph. Thus, training from scratch (which was necessary for the network evaluation) is no longer needed, and the tensor of shared weights can be finetuned via gradient updates in-between controller updates. While weight sharing technique allows for significant reduction in computations, it also tends to distort the performance ranking of intermediate candidate networks, leading to potentially inferior models being chosen \cite{Sciuto2019}.

Evolutionary algorithms arose once again as a primary competitor to reinforcement learning based solutions. CoDeepNEAT \cite{Miikkulainen2017} adapts the well-known NEAT procedure for deep networks by using two separate populations - \emph{blueprints} and \emph{modules} - for easier representation of repeating patterns. Genetic CNN \cite{Xie2017} encodes layer connectivity in a population of binary strings and runs a standard genetic algorithm. EvoCNN \cite{Sun2019a} instead opts for variable-length gene encoding to represent networks of arbitrary depth, where crossover is made possible via matching the genes that share a type (e.g. a pooling gene can only be matched with another pooling gene). Real \textit{et~al.} \cite{Real2017} run a distributed large-scale evolutionary process directly on a population of networks, where mutations can alter the network structure, parameters, or training process. The following work of Real \textit{et~al.} \cite{Real2018a} combines the evolutionary approach with the NAS-Net search space, surpassing reinforcement learning in anytime accuracy and setting a new state-of-the-art performance on the popular CIFAR-10 dataset, as well as generating comparatively simpler models. However, the computational cost remains extensive, clocking above 3000 GPU days. The similar approach is taken by the automatically evolving CNN (AE-CNN) \cite{Sun2019}, which runs a genetic algorithm on the population of networks composed of customized ResNet and DenseNet blocks, achieving competitive results.

While deep networks can be difficult for the neuroevolution to handle, a viable alternative can be found in expanding the operation set of the shallow networks, allowing for more powerful representations. Generalized Operational Perceptron (GOP) \cite{Kiranyaz2017} model substitutes the standard neuron by offering a wider choice of nodal and pooling operations instead of the standard multiplication and addition. The choice of operations can be optimized simultaneously with the network architecture by a greedy incremental procedure. Operational Neural Networks (ONNs) \cite{Kiranyaz2019}, composed of such units, have been shown to achieve superior performance to CNNs on some practical problems. Most recently heterogeneous GOP structures, where each layer can have neurons with differing operations, have received increasing attention \cite{Tran2019}. While flexibility of operators allows ONNs to stay relatively shallow, it also results in a vast unstructured design space which is computationally costly to traverse.

Many recent works in architecture optimization utilize various techniques to reduce the computation needed, primarily by simplifying the evaluation procedure. SMASH \cite{Brock2017} learns a hypernetwork that can predict weights for all the connections of an arbitrary deep network (given a specific representation), which reduces the need for training and makes random search a viable solution for discovering architectures. Progressive Neural Architecture Search (PNAS) \cite{Liu2018} uses a separate recurrent network to approximately rank the candidate models without training them, allowing the search to focus only on more promising options. NASH \cite{Elsken2018b} and LEMONADE \cite{Elsken2019} take advantage of network morphisms---operations that modify the structure of a trained network without affecting its output---to navigate the search space without training the models from scratch. Differentiable Architecture Search (DARTS) \cite{Liu2018a} provides a continuous relaxation of the NAS-Net cell structure problem and performs the search via gradient descent, iterating between the architecture and weight updates. While relatively more efficient in terms of computation, these methods do not address the issues of interpretability and semantic complexity.

There exists a number of works that model the network construction as a probabilistic process, sharing some similarities with the proposed approach. Methods based on reinforcement learning, such as NAS and its successors, use the probability of a given network to be produced from the current policy as a weight for the corresponding reward. InstaNAS \cite{Cheng2018} also has reinforcement learning at its core, but differs from other algorithms in this group, as it takes an instance-aware approach. Specifically, InstaNAS processes each data point by a separate network (a path within a large trained model), sampled from a parameterized distribution. NASBOT \cite{Kandasamy2018} models the architecture search as a Gaussian process. To facilitate this, the authors introduce a (pseudo)distance in the network design space and utilize an evolutionary algorithm as an optimizer.  The most similar approach to ours is Probabilistic Neural Architecture Search (PARSEC) \cite{Casale2019a}. As in our work, PARSEC explicitly models a structural distribution of the neural networks, including the assumption of independence between individual operations. However, this distribution operates on a level of NAS-Net cell, while our method models the network as a whole without forced repeating patterns. Moreover, the search procedure is different: PARSEC uses Monte Carlo empirical Bayes to iteratively update both the architectural priors and the tensor of shared weights, while we completely recompute the marginal probabilities over a subset of the samples and do not use weight sharing.

Our work draws inspiration from the estimation of distribution algorithms (EDAs) -- the family of optimization methods originating from mid-1990s, which are closely related to genetic algorithms \cite{Pelikan2002}. While most evolutionary algorithms maintain a candidate population, which implicitly defines the probability distribution of the solutions, EDAs define this distribution explicitly and tune its parameters throughout the optimization process.  Our work mainly draws on two discrete univariate EDAs, Population-Based Incremental Learning (PBIL) \cite{Baluja1994} and Univariate Marginal Distribution Algorithm (UMDA) \cite{Muhlenbein1997}. PBIL generates an intermediate population via sampling, applies a selection procedure, and updates the probabilistic model in the direction of selected samples, using a learning rate parameter. UMDA maintains the population of solutions, estimates a set of marginal probabilities from the best candidate(s), and uses them to produce the population of the next generation. For more information on EDAs, their applications and recent developments, we direct the reader to the survey by Hauschild and Pelikan \cite{Hauschild2011}. To the best of our knowledge, ours is the first work to explicitly apply the EDA formulation to the deep neural network architecture search problem.

\section{Methodology}\label{method}
In this section, we describe and justify the proposed network representation, the design of the proposed algorithm Architecture Search by Estimation of network structure Distributions (ASED), as well as additional techniques to improve its capabilities.
\subsection{Search Space and Network Representation}
The problem of optimizing the structure of a neural network is extremely high-dimensional. The choice of layer types (convolution, pooling) alone produces a combinatorial problem that grows exponentially with the increase in depth, and that is without taking into account layer hyperparameters (filter size, stride, channel count) and weights. Connectivity patterns add another dimension of complexity, as structures such as skip connections and parallel branches have been found beneficial in manually designed models \cite{Huang2017a}. For this reason many recent architecture optimization algorithms, starting with NAS-Net by Zoph \textit{et~al.} \cite{Zoph2017}, utilize a constrained search space based on repeated structural motifs. Instead of searching for the architecture of the entire network, they instead work with \emph{cells}, which are small subnetworks containing only a few layers. The target network is then constructed by repeating the cell a given number of times. This relaxation allows the cells to have almost arbitrary structures while maintaining the viability of the search. Another advantage is the directly controllable trade-off between the network power and complexity by varying the number of cell repetitions. It is common to speed up the search by using less cells and then increase their number for the final evaluation of the discovered architecture. However, the unavoidable natural drawback of this approach is the fact that only a small subset of network design space is reachable with such constraints, and potentially better architectures may not be discoverable. Therefore, we opt for optimizing the whole network simultaneously.

We model a deep neural network as a multivariate random variable coming from a known probability distribution. For the sake of tractability, we consider only the choices of layer types for optimization, resulting in a discrete distribution, while other hyperparameters are not directly tuned by the search procedure. Specifically, we bind the values of filter sizes and strides with the layer type choices and set the channel count to an externally defined constant for all the layers. We denote the set of possible layer types as $L$ and call it \emph{the layer library}. For the purpose of this work, we include the following ten common operations in the library (each with the corresponding shorthand notation): 
\begin{itemize}
    \item [id] identity (output is equal to input),
    \item [c1] 1x1 convolution,
    \item [c3] 3x3 convolution,
    \item [c5] 5x5 convolution,
    \item [c7] 7x7 convolution,
    \item [d3] 3x3 dilated convolution (with dilation rate of 2),
    \item [d5] 5x5 dilated convolution (with dilation rate of 2),
    \item [m2] 2x2 max pooling,
    \item [m3] 3x3 max pooling (with stride 2), 
    \item [a3] 3x3 average pooling (with stride 2).
\end{itemize}

We assume that the choice of each layer in a CNN is independently distributed. While this assumption is unlikely to hold in practice, it simplifies the formulation, and inter-layer interactions are implicitly taken into account during the search. Multivariate generalizations of the proposed method can potentially offer improvements and are a promising future work direction. Given our assumption, a discrete distribution of network structures can be represented as a matrix of probabilities $P$, where each row describes a layer and $P_{ij} \in [0,1]$ is the probability of $i$-th layer being the $j$-th layer type from $L$. Matrix $P$ is henceforth called \emph{prototype}. The dimensions of $P$ are $N \times |L|$, where $N$ is the current number of layers in the network.

The probabilistic representation has a number of inherent advantages over maintaining the population of networks. Matrices have a wide range of available optimization approaches; many existing optimization heuristics outside of the scope of this work are straightforwardly applicable to the proposed representation. The prototype offers intuitive insight into the anytime state of the search, as the probability mass is always explicitly assigned for every point of the design space. The convergence of the search is easy to determine by how close the layer probabilities are to their extremes. Finally, the proposed representation can offer significant implementation advantages in a distributed setting, as only a small prototype matrix needs to be transferred between computational nodes, rather than full-scale models.

Evaluation of the prototype involves sampling network structures from it and measuring their performance (and potentially other metrics) on the given problem. By default, every sampled network has to be trained from scratch, which can result in fairly high computational costs, especially for deep prototypes (with large values of $N$). Weight sharing, when candidate networks reuse trained weights from the previous search iterations, is commonly used in other methods to reduce the overall training time. A form of weight sharing can be implemented in our framework by storing every newly encountered structure with its performance in a dictionary and retrieving these values instead of retraining, when such structure is sampled again. However, this solution has drawbacks of its own: as we continually deepen the prototype, the chances of repeatedly sampling the same structure are actually very low, while maintaining the dictionary would noticeably increase the memory requirements. In addition, weight sharing can give an unfair advantage to inferior structures, thus damaging the overall search \cite{Sciuto2019}. For these reasons we opt to contain the computational costs by other means: specifically, limiting the value of $N$ and reducing the number of epochs for candidate training. We consider some other speedup options in the analysis section.

\subsection{Search Algorithm}
To construct an iterative architecture search algorithm with the above representation, three elements need to be defined - initialization, update and stopping condition. The proposed algorithm, denoted ASED, operates on a single prototype for the sake of simplicity. The depth of all networks on a given search step is the same due to the fixed prototype dimensions; to search across architectures of different sizes, we gradually increase the depth after each update step. While the prototype rows are never removed, the inclusion of identity in our layer library means that, in practice, networks with less than $N$ layers can be represented and discovered at any search stage.

The prototype $P$ is initialized as a $N_{init} \times |L|$ matrix with every element set to $1/|L|$, where $N_{init}$ is a starting layer count. While a more specific prior can be given, the uniform initial distribution ensures that every reachable architecture is equally likely to be considered, which helps to emphasize early exploration. The choice of $N_{init}$ should be carefully considered, as a small value can result in premature convergence without sufficiently exploring the larger portion of the design space, but a large value can cause the search to be "lost" unless an impractically large number of samples is evaluated (due to the curse of dimensionality).

To update the prototype, sampling of $K$ candidate networks is performed first, with each layer independently selected from the discrete distribution given by the corresponding row of the prototype matrix. Each candidate model is then trained and evaluated on the target problem, and the temporary population is sorted by validation performance. While we evaluate and track the classification accuracy, we opt for another, additional measure to compute the candidate ranking. We adopt the multi-class Matthews coefficient \cite{Gorodkin2004}, which is designed to be robust to the class imbalance in the data, allowing the search to operate reliably in such cases. The formula of the multi-class Matthews coefficient is as follows:
\begin{equation}\label{eqM}
    m = \frac{\sum\limits_{klm} (C_{kk} C_{lm} - C_{kl} C_{mk})}{\sqrt{\sum\limits_k (\sum\limits_l C_{kl})(\sum\limits_{\mathclap{\substack{l' \\ k' \neq k}}} C_{k'l'})} \sqrt{\sum\limits_k (\sum\limits_l C_{lk})(\sum\limits_{\mathclap{\substack{l' \\ k' \neq k}}} C_{l'k'})}},
\end{equation}
where $C$ is a confusion matrix. The value of the multi-class Matthews coefficient ranges between a data-dependent negative value ($\geq -1$) and $+1$. Whenever the denominator of the fraction in Equation \ref{eqM} is zero, we set the output value to the lowest possible: -1.

Given the ranking, the best $K_s < K$ models are then selected to directly induce the new prototype, which, due to the independence assumption, takes the following form:
\begin{equation}\label{eq1}
    P_{ij} = \frac{1}{|K_s|}\sum_{k=1}^{K_s} x_{kij},
\end{equation}
where $x_{kij}$ is an indicator variable that is equal to 1 if $k$\nobreakdash-th selected candidate network has $j$\nobreakdash-th library item as $i$\nobreakdash-th layer, and 0 otherwise. This update step is equivalent to the one used in the UMDA algorithm \cite{Muhlenbein1997}. Every update is followed by an addition of one or more rows to the prototype, according to the predefined schedule (denoted $n(t)$). These new layers are initialized with a uniform distribution. Note that we avoid explicitly preserving the best candidates between updates (the technique known as elitism). As our approach starts from the solution space of low dimensionality and gradually increases it, the bias towards early dominant solutions will result in premature convergence.
The complete description of the ASED procedure is given in Algorithm \ref{algo1}. The search stops when the specified iteration limit $t_{max}$ is reached or all the values of the prototype matrix become strictly 0 or 1, which indicates complete convergence. A single network architecture with the highest probability is selected to be the final output.
\begin{figure}[!t]
    \begin{algorithm}[H]
    \caption{Architecture Search by Estimation of Network Structure Distribution (ASED)}\label{algo1}
        \begin{algorithmic}[1]
            \State \textbf{Input:} $L$, $N_{init}$, $t_{max}$, $K$, $K_s$, $n(t)$
            \State $N \gets N_{init}$
            \For{$i \in \{1, \dots, N_{init}\}$, $j \in \{1, \dots, |L|\}$}
            \State $P_{ij} \gets 1/|L|$
            \EndFor
            \For{$t \in \{1, \dots, t_{max}\}$}
                \State Sample $K$ candidate networks from $P$
                \State Train and evaluate candidate networks
                \State Sort candidate networks by validation performance
                \State $S \gets K_s$ best performing candidate networks
                \State Recompute $P$ based on $S$ (Eq.~\ref{eq1})
                \State Add $n(t)$ new rows to $P$
                \For{$i \in \{N+1, \dots, N+n(t)\}$, $j \in \{1, \dots, |L|\}$}
                    \State $P_{ij} \gets 1/|L|$
                \EndFor
                \State $N \gets N+n(t)$
                \If{$\forall i,j$ $P_{ij} \in \{0,1\}$}
                    \State \textbf{break}
                \EndIf
            \EndFor
            \State \textbf{return} $P$
        \end{algorithmic}
    \end{algorithm}
\end{figure}

\subsection{Convergence Control Techniques}
While the described search procedure navigates the search space by progressively narrowing down the region under consideration and should be capable of avoiding local optima, it can still get stuck in a local optimum and hence exhibit premature convergence. As the search progresses, individual layer type probabilities tend to approach either 0 or 1 regardless of their immediate impact on the network performance, as is established in the theory of EDAs \cite{Zhang2004a, Friedrich2016}. The proposed algorithm does not allow for any mechanisms to limit this; in fact, such an effect is desirable for the search convergence. Moreover, once a probability has achieved the value of exactly 0 or 1, it becomes fixed and will not change thereafter, as all of the sampled networks will be the same with respect to the type of the corresponding layer. In case of 0 the corresponding operation is no longer considered, while in case of 1 the layer choice is made permanent, meaning that the dimensionality of the problem is essentially reduced from that point on. A subset of network structures becomes unreachable, which can be beneficial for navigating the design space, but can also mean the loss of potentially better solutions. We consider two different techniques to address this issue.

A common technique in EDAs involves capping the probabilities, such that extreme values are not achievable and each element instead spans the predefined range $[p_{min},p_{max}]$, where $p_{min} > 0$, $p_{max} < 1$. In our setting, this means that there is always at least the probability of $p_{min}$ for each layer type to be selected in any position, removing irreversible choices. Probability capping is implemented by simple row-wise proportional normalization of the prototype matrix after every prototype update step. We adopt the approach where the upper cap $p_{max}$ is explicitly given as a parameter and the lower cap is then computed as
\begin{equation}
\label{eq_pmin}
    p_{min} = \frac{1-p_{max}}{|L|-1}.
\end{equation}

The normalization itself is then performed as follows:
\begin{align}
\label{eq_norm}
    & p'_{ij} =
    \begin{cases}{}
        p_{min} & \text{if } p_{ij} \leq p_{min} \\
        p_{ij} \cdot m_i & \text{if } p_{ij} > p_{min}
    \end{cases} \nonumber \\
    & m_i = \frac{\sum B_i + \sum S_i - |S_i| \cdot p_{min}}{\sum B_i},
\end{align}
where $S_i = \{ p_{ij} | p_{ij} \leq p_{min} \}$ and $B_i = \{ p_{ij} | p_{ij} > p_{min} \}$ for $i \in \{1, \dots, N \}$. Note that $B_i$ cannot be 0 as the layer probabilities sum to 1.

Another way to prevent the search from prematurely converging is to additionally modify the prototype between iterations. One can, for instance, apply a small random perturbation, similar to how mutation is used in evolutionary algorithms. However, due to the search being driven by sampling, the effect of such mutation would be either insignificant or highly unpredictable. Instead, we opt for another operation, which we call \emph{prototype inversion}, that replaces high probabilities with low values and vice versa. This prompts the search to explore exactly the previously discarded regions of the search space, while the currently dominant choices become extremely unlikely (the latter aspect evokes similarities to the well-known tabu search, which is an optimization technique that explicitly forbids the reuse of already seen solutions \cite{Glover1998}). Naturally, such an inversion operation is highly destructive and can prevent the search from progressing, so it needs to be executed only at some iterations of the algorithm. Additionally, we save the current prototype just before inverting it to make sure the information is not lost, which essentially means the ASED algorithm can produce multiple solutions before the stopping condition is met.

To establish an inversion condition, we need to find the measure of convergence, as performing the inversion too early and/or too often can hinder the search process. $L_2$\nobreakdash-norm of the probability vectors is suitable for this purpose, as it spans the interval $[1/\sqrt{|L|},1]$. Here, the lower bound corresponds to the uniform distribution and the upper bound is achievable only when a single element (layer type) takes value 1 with every other being 0. The $L_2$\nobreakdash-norm of each prototype row is thus a measure of the certainty of the layer choice and increases as the search progresses. The condition for triggering the prototype inversion can then be a threshold on the $L_2$\nobreakdash-norm of the prototype, averaged over all the rows (as they are assumed independent). If the inversion is used, this condition is checked at every iteration after the prototype is updated. The newly added uniformly distributed rows are ignored for the purposes of the mean norm calculation.

The inversion operation is implemented by subtracting each probability value from 1 (e.g., 0.85 becomes 0.15). For probabilities that have taken values of 0 or 1 this operation does not have the intended effect, as they still limit the exploration space. To suppress these extreme values, the inversion is followed by the probability capping normalization defined in Eq.~(\ref{eq_norm}), using the same parameters $p_{max}$ and $p_{min}$. As no probability goes to zero as a result of inversion, previously reachable solutions remain reachable, allowing for potential backtracking. We consider two types of inversion operation: the \emph{full inversion} and the \emph{partial inversion}. The former is applied row by row to the whole prototype. The latter is less destructive as it only applies to the subset of prototype rows which have the highest $L_2$-norm. The specific number of such rows is empirically set to $\lfloor\sqrt{N}\rfloor$. The partial inversion thus applies only to some of the most converged layers, preserving less confident choices as they are.

As both described techniques are simple mathematical operations on the prototype matrices, they do not incur significant computational costs by themselves. However, as they influence the search behaviour of the algorithm towards slowing down the convergence, more iterations may be required to reach the solution of the same level of complexity as with the baseline variant. With respect to the given stopping condition, the proposed techniques can indirectly increase the overall running time of the algorithm, although the specific impact can be evaluated only empirically on a case-by-case basis.

\subsection{Non-Linear Layer Connectivity}
The default prototype specification supports only the simplest architectures, where the flow of the input data through the layers is strictly sequential. However, connections between non-adjacent layers, also known as \emph{shortcuts}, play an important role in the powerful CNN models, such as ResNet \cite{He2015} and DenseNet \cite{Huang2017a}. Shortcuts counteract the problem of vanishing gradients by ensuring the efficient flow of the backwards propagated signal, which allows models of much larger depth to be reliably trained. While we constrain layer counts during the architecture search due to computational limitations, the use of shortcuts still simplifies and speeds up the training of candidate models as the depth increases. By making deeper models more competitive against shallow ones already in the early training stages, shortcuts discourage premature convergence of the prototype. Thus, while implicit, they grant similar benefits as the techniques described in the previous subsection. Of course, our framework also benefits from shortcuts via increased expressiveness and generality. 

There are several options for introducing shortcuts to the proposed formulation. One possibility is to define another prototype matrix, which would store the connectivity patterns, i.e., presence or absence of a connection, between layers. This additional prototype can be operated on with any procedure suitable for the original, such as the one given in Algorithm \ref{algo1}. The two matrices can be concatenated and optimized jointly, or they can be iterated between, introducing an additional step to the search procedure. It is worth noting, however, that such an approach significantly increases the problem complexity. Additional dimensions of the search space would require either exponentially more samples (in the joint case) or exponentially more iterations (in the iterative case) to ensure sufficient exploration. As a compromise between complexity and expressivity, we consider another approach --- \emph{fixed shortcut patterns}. We design simple shortcut-generating rules, inspired by existing deep CNN models, and apply them to any sampled candidates throughout the search, as well as to the ultimately discovered architectures (while ensuring the solutions remain valid). As the rule does not change over time, the search procedure can be expected to select structures which take the most advantage of the given shortcut pattern. This simple approach does not incur any additional computational costs. In fact, it can potentially speed up the training of the deeper networks, but this is not currently taken advantage of, as we use fixed training schedules for simplicity. 

\begin{figure}[!ht]
\centering
\includegraphics[width=\linewidth]{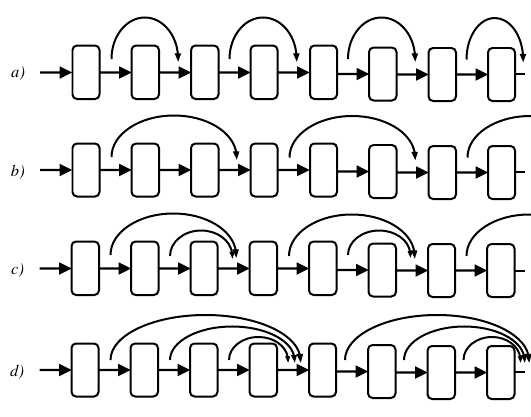}
\centering
\caption{Illustrated shortcut patterns: a)~both, $D=1$; b)~residual, $D=2$; c)~semi-dense, $D=2$; d)~semi-dense, $D=3$.}
\label{fig_sh_illust}
\end{figure}

We consider two types of shortcut patterns --- the \emph{residual pattern} and the \emph{semi-dense pattern}. Both can be characterized by a single parameter $D\geq1$. \emph{Residual pattern} is inspired by the residual connections of ResNet \cite{He2015}. In this case, the shortcut connects the output of the current layer to the output of the layer which is $D$ positions after (e.g., the following layer if $D=1$). Residual shortcuts cannot intersect or overlap, i.e., there can be no starting point between another starting point and its corresponding endpoint. In the case of the \emph{semi-dense pattern}, every group of $D$ consequent layers have their outputs connected to the single endpoint at the output of the ($D+1$)th layer. It is inspired by the DenseNet \cite{Huang2017a} with some simplification (the original pattern would have all the layers within the group connected). These shortcut patterns are illustrated in Fig.~\ref{fig_sh_illust}.

As in other deep models, the signal arriving via a shortcut is combined with the primary signal at the endpoint by simple addition. As in the early ResNet, we follow the post-activation pattern, i.e., the activation function is applied after signals are combined (but batch normalization, if applicable, precedes the signal combination). A notable issue for the proposed approach is the compatibility of the signal dimensions. In handcrafted architectures, shortcuts are typically applied within layer blocks that do not perform downsampling, and, therefore, the shapes of the primary and the shortcut signals always match. This does not necessarily hold within our framework: while convolutions are zero-padded to preserve dimensions between input and output, any layer can potentially be a pooling layer, breaking the compatibility. We resolve this issue by keeping track of the pooling operations that the primary signal is undergoing and applying them to the shortcut signal before adding the two together. Compatibility with regards to the channel number is guaranteed, since it remains constant throughout the network, with the exception of the original input, which is not allowed to be the starting point of a shortcut. It is also worth noting that, since our framework allows every layer to potentially represent an identity function, the span of some of the shortcuts may in practice be less than $D$. 

\section{Experimental Results and Analysis}
In this section, we describe the experimental setting and obtained results and discuss their implications for this and future work.
\subsection{Experimental Setting}\label{setting}
Following the previous works of the field, the proposed ASED algorithm is validated in the image classification setting. We consider two datasets of differing difficulty and scale. As a simpler problem we utilize USPS dataset \cite{Hull1994}, which contains grayscale samples of handwritten digits of 16x16 pixels each. USPS has 10 classes, 7291 training examples, and 2007 test examples. The larger problem is CIFAR-100 dataset \cite{krizhevsky2009learning}, which is one of the standard NAS benchmarks. It consists of 3-channel 32x32 RGB images. It contains 50K training examples and 10K test examples of 100 different classes. To allow for evaluation of candidate networks without leaking the test data, for both datasets we sample 20\% of the original training images, maintaining class balance, to obtain validation sets. We use the original test sets only to report the performance of the final discovered model. Classification accuracy is used as the performance metric. We do not use any preprocessing for USPS. For CIFAR-100 the standard preprocessing procedure is applied: images are zero-padded by 4 pixels on each side, followed by random cropping down to 32x32 and horizontal flipping with probability 0.5, as well as normalization to zero mean and unit variance. We do not employ other regularization techniques, such as cutout, drop-path, or shake-shake.

The algorithm parameters are set as follows. The search is initialized by sampling 10K networks from the uniform 5-layer prototype, which covers 10\% of the design space (as the current library permits $10^5$ possible 5-layer networks). At every iteration of the search, $K=1000$ networks are sampled, trained, and ranked, with the top $K_s=100$ forming the next prototype. As adding layers is initially affordable, but becomes more expensive later, the following growth schedule is adopted:
\begin{equation}
    n_t =
    \begin{cases}
    2 & \text{if } t \in \{1, 2\} \\
    1 & \text{otherwise.}
    \end{cases} 
\end{equation}

For USPS dataset we additionally consider a modified set of settings to start from smaller networks, which allows for more gradual navigation of the search space in the early stages. With the modified search settings a 2-layer uniform prototype is used as a starting point (instead of a 5-layer one). The sampling between iterations is also reduced: 100 networks are generated and top 10 are selected to induce the next prototype. This results in the smaller number of candidates being evaluated, speeding up the search by approximately a factor of ten.

During the search, the channel count of all convolutional layers is set to 32. To avoid the mismatch of signal dimensions, every convolutional layer has its output padded to match the input; therefore, only the pooling layers can perform downsampling. We use PReLu as an activation function and apply the corresponding initialization policy of He \textit{et~al.} \cite{He2015a}. To make the discovered architectures output the class predictions, we perform global average pooling after the last sampled layer, followed by a fully connected layer of 100 PReLu-activated neurons, dropout with rate 0.5, and a softmax layer. Batch normalization is not used within the candidate networks during the search; however, the final discovered architecture has batch normalization applied after every convolutional layer to maximize evaluation performance.

Training numerous deep networks from scratch incurs the majority of computational expenses of the search procedure. We therefore use a different, less intensive training regime, denoted as \emph{brief training}, to produce metrics for candidate ranking during the search. \emph{Full training} is reserved only for the evaluation of the final solution discovered by the architecture search. Both settings use stochastic gradient descent (SGD) with momentum of 0.9 to minimize the cross entropy loss. Brief training runs for 20 epochs, using the learning rate of $10^{-2}$ for the first 10 epochs and $10^{-3}$ thereafter. Full training runs for 200 epochs and uses the following schedule of learning rates: 0.01 for epochs 1-60, 0.02 for epochs 61-120, 0.004 for epochs 121-160, and 0.001 for epochs 161-200. Full training also uses $L_2$-norm weight regularization with the coefficient of $10^{-4}$ (excluding PReLU weights, as recommended in \cite{He2015a}) and imposes the maximum $L_2$-norm constraint of 0.5 on weights. The batch size is set to 64 for USPS and 128 for CIFAR-100.

The settings for convergence control and non-linear connectivity variants are given as follows. For the variants that use probability capping we set $p_{max}$ to 0.9. The $L_2$-norm threshold for both full inversion and partial inversion is set to 0.65, as that corresponds to the middle of the interval of the possible values. For the evaluation of shortcut-generating rules we consider $D=2$ and $D=3$, as $D=1$ has limited impact on the gradient flow and $D>3$ results in too few shortcuts created, given the network depth limitation. We do not consider the convergence control and the shortcut patterns jointly: since both of them act like regularizers on the search procedure, their significance is easier to analyze separately. The total number of iterations is $t_{max}=9$ for the baseline and probability capped variants, which, under the adopted schedule, corresponds to the maximum layer count of 16. However, for the inversion experiments the search is run further to allow for observing the results after multiple instances of inversion triggering. The achievable layer count is 22 for these variants.

All of our experiments are implemented in PyTorch and performed on a workstation with 4 GeForce 1080Ti GPU units. The running times are included with the reported results. The source code is available on GitHub \footnote{~\url{https://github.com/anton-muravev/ased}}.

\subsection{Architecture Search Performance on USPS}

\begin{figure*}[!ht]
\centering
\includegraphics[trim=0 0 150 0]{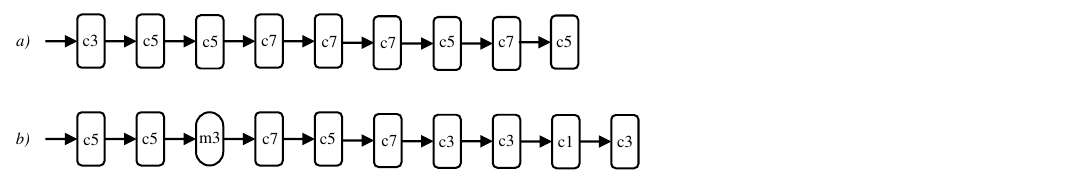}
\centering
\caption{Best discovered structures on USPS for a)~default settings, b)~modified settings. Shorthand notation from Section~\ref{method}. Pooling operations denoted by oval shapes.}
\label{fig_best_nets_usps}
\end{figure*}

\begin{table*}[!ht]
\renewcommand{\arraystretch}{1.3}
\caption{Comparison of the error rates (full training mode) on USPS dataset}
\label{table_usps}
\centering
\begin{tabular}{|c|c|c|c|c|}
\hline
Model & Test error (\%) & Parameter count & Model depth & Search time (GPU hours)\\
\hline\hline
Maxout S-ResNet+ENR+SVM \cite{Oyedotun2018} & 2.34 & 169K & 54 & N/A\\
\hline
Maxout S-ResNet+ENR+FS+SVM \cite{Oyedotun2018} & 2.19 & 169K & 54 & N/A\\
\hline\hline
ASED default, 16 channels & 2.49 & 77K & 9 & 64.0\\
\hline
ASED default, 32 channels & 2.25 & 305K & 9 & 64.0\\
\hline
ASED modified, 8 channels & 2.64 & 12K & 10 & 6.4\\
\hline
ASED modified, 16 channels & 2.25 & 46K & 10 & 6.4\\
\hline
ASED modified, 32 channels & 2.25 & 182K & 10 & 6.4\\
\hline
\end{tabular}
\end{table*}

To the best of our knowledge, the current state of the art results on USPS are achieved by Oyedotun et. al. \cite{Oyedotun2018}. Their solution is based on a well-known ResNet architecture, modified by a number of techniques: maxout activation function, elastic net regularization, and feature standardization. Furthermore, the final classification is obtained via an SVM that is trained on the features from the final convolutional layer. We compare the reported results from \cite{Oyedotun2018} with the solutions discovered by the baseline ASED in its default and modified configurations. The comparison between ResNet and both ASED setups is given in Table~\ref{table_usps}. The discovered architectures are shown in Fig.~\ref{fig_best_nets_usps}, using the notation defined in Section~\ref{method}.

The networks discovered by ASED clearly produce competitive results to modified S-ResNet. While the default search configuration shows a clear tradeoff between classification accuracy and the number of parameters, the modified search configuration can produce very competitive architectures with just 16 channels. The discovered architecture is significantly smaller than S-ResNet in both depth and overall memory requirement, while achieving a similar test error rate. ASED is, therefore, capable of discovering very efficient architectures in smaller problem domains. The discovered architecture notably differs from common patterns of handcrafted designs, lacking pronounced block-like repeating patterns and taking advantage of convolutions with larger kernels. This shows the potential of the neural architecture search to automatically discover novel and potentially superior network structures, which may lie outside the standard design spaces while not being discoverable by cell-based architecture search.

\subsection{Architecture Search Performance on CIFAR-100}

We report the results of running the proposed ASED algorithm on CIFAR-100 with the default search configuration described above. In addition to the baseline given by Alg.~\ref{algo1}, we also evaluate the proposed modifications of the search: probability capping (denoted ProbCap for clarity), the full inversion variant, the partial inversion variant, and the shortcut-generating rules of both types. 

\begin{figure*}[!ht]
\centering
\includegraphics[width=\textwidth]{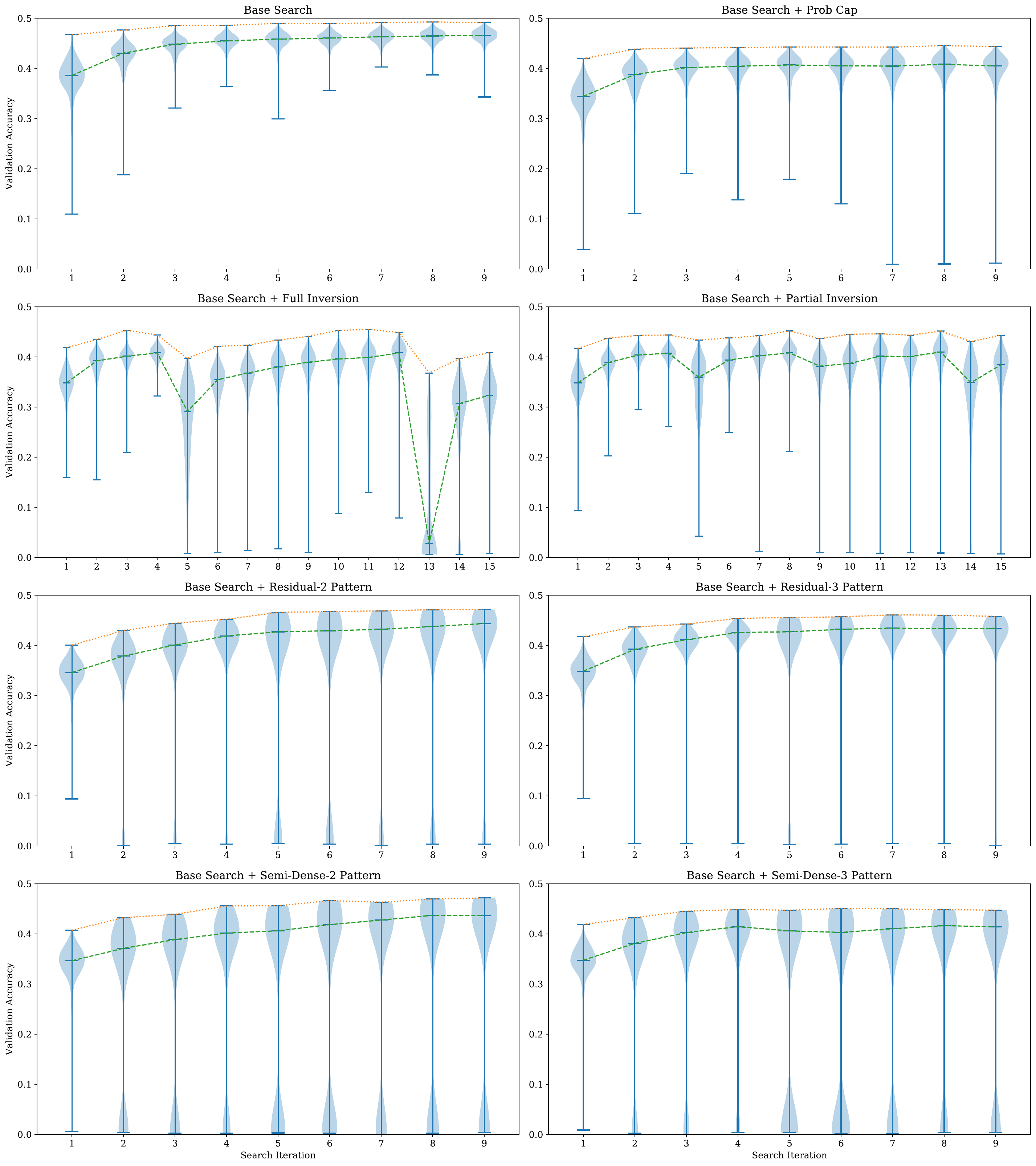}
\centering
\caption{Distributions of validation accuracies (brief training mode) on CIFAR-100 across search iterations. The line plots show the maximum and median accuracies of the sampled networks.}
\label{fig_acc}
\end{figure*}

Fig.~\ref{fig_acc} shows the distributions of candidate models' performances on the validation set throughout the search iterations, under the brief training regime. The maximum and median accuracies (shown by the red dotted and green dashed lines respectively) show steady convergence patterns in most cases. The inversion variants are notable exceptions, as the inversion operations result in clearly noticeable, although temporary, accuracy drops. Probability capping stands out as an example of the premature convergence and stagnation, where the performance does not reach the levels of other ASED variants. It is also worth noting that the networks with semi-dense connection patterns appear to be more sensitive to the layer type changes, as throughout the search process a notable portion of the sampled structures continues to exhibit poor performance. 

\begin{figure*}[!ht]
\centering
\includegraphics[width=\textwidth]{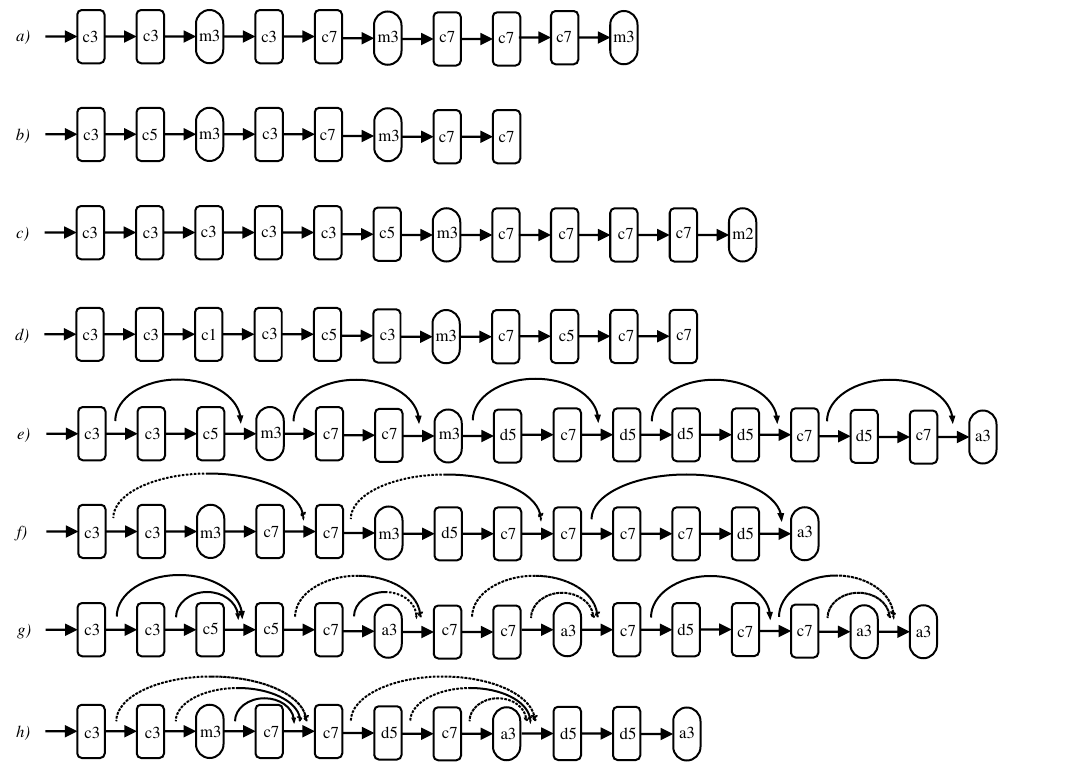}
\centering
\caption{Best discovered structures for a)~base procedure, b)~probability capping, c)~full inversion, d)~partial inversion, e)~residual-2 pattern, f)~residual-3 pattern, g)~semi-dense-2 pattern, h)~semi-dense-3 pattern. Shorthand notation from Section~\ref{method}. Pooling operations denoted by oval shapes. Dotted lines indicate skip connections with pooling.}
\label{fig_best_nets}
\end{figure*}

While limiting the number of training epochs does allow for much faster evaluations during the search process, the performance estimates obtained this way (shown on Fig.~\ref{fig_acc}) are pessimistic and not fully representative of the underlying models' capabilities. Therefore, we also conduct an extended evaluation of the best discovered networks by adding batch normalization modules and using the full training regime (see Section~\ref{setting}). In this comparison, we also include the best architecture from the initialization sample (generated from the 5-layer uniform prototype), as well as the best 16-layer architecture from 1000 networks generated from a uniformly random prototype. Every configuration is trained from scratch with the convolution channel counts of 32, 64, 128, and 256. The results are reported in Table~\ref{table_own_comparison}. The indicated layer count excludes any identity layers. The best-performing network structures from each algorithm variant are shown in Fig.~\ref{fig_best_nets}.

\begin{table*}[!ht]
\renewcommand{\arraystretch}{1.3}
\caption{Comparison of best discovered architectures by their test accuracy (full training mode) on CIFAR-100 dataset}
\label{table_own_comparison}
\centering
\begin{tabular}{|c|c|c|c|c|c|c|c|c|c|}
\hline
\multirow{2}{2em}{Model Source} & \multirow{2}{2em}{Layer Count} & \multicolumn{2}{c}{32 channels} & \multicolumn{2}{|c|}{64 channels} & \multicolumn{2}{c}{128 channels} & \multicolumn{2}{|c|}{256 channels} \\
\cline{3-10}
& & Acc. & Par. & Acc. & Par. & Acc. & Par. & Acc. & Par. \\
\hline\hline
Initialization & 5 & 0.4898 & 131K & 0.5835 & 513K & 0.6419 & 2.0M & 0.6846 & 8.1M \\
\hline
Random uniform & 15 & 0.5794 & 223K & 0.6621 & 879K & 0.7200 & 3.5M & 0.7499 & 13.9M \\
\hline
ASED & 10 & 0.5659 & 224K & 0.6582 & 886K & 0.7102 & 3.5M & 0.7483 & 14.1M \\
\hline
ASED + Prob Cap & 8 & 0.5483 & 190K & 0.6330 & 751K & 0.6963 & 3.0M & 0.7442 & 11.9M \\
\hline
ASED + Full Inversion & 12 & 0.5827 & 268K & 0.6728 & 1.1M & 0.7297 & 4.2M & \bfseries 0.7729 & 16.9M \\
\hline
ASED + Partial Inversion & 11 & 0.5748 & 236K & 0.6641 & 932K & 0.7249 & 3.7M & 0.7652 & 14.8M \\
\hline
ASED + Residual-2 & 16 & 0.6396 & 419K & 0.6872 & 1.7M & 0.7282 & 6.6M & 0.7485 & 26.5M \\
\hline
ASED + Residual-3 & 13 & 0.6194 & 367K & 0.6689 & 1.5M & 0.7068 & 5.8M & 0.7315 & 23.2M \\
\hline
ASED + Dense-2 & 15 & \bfseries 0.6461 & 392K & \bfseries 0.6984 & 1.6M & \bfseries 0.7398 & 6.2M & 0.7610 & 24.8M \\
\hline
ASED + Dense-3 & 11 & 0.6092 & 251K & 0.6678 & 1.0M & 0.7084 & 3.9M & 0.7348 & 15.8M \\
\hline
\end{tabular}
\end{table*}

\begin{table*}[!ht]
\renewcommand{\arraystretch}{1.3}
\caption{Test performance comparison of established and automatically discovered architectures on CIFAR-100 dataset}
\label{table_cifar100_global}
\centering
\begin{tabular}{|c|c|c|c|c|}
\hline
Method & Accuracy (\%) & Parameter count & Model depth & Search cost (GPU days)\\
\hline\hline
FractalNet \cite{Larsson2017} & 76.7 & 38.6M & 21 & N/A \\
\hline
Shake-Shake \cite{Gastaldi2017a} & \bfseries 84.2 & 26.2M & 26 & N/A \\
\hline
Wide ResNet 28-10 \cite{Zagoruyko2016} & 80.4 & 36.5M & 28 & N/A \\
\hline
DenseNet-BC \cite{Huang2017a} & 82.8 & 25.6M & 190 & N/A \\ 
\hline\hline
Genetic CNN \cite{Xie2017} & 70.9 & -- & 17 & 17 \\
\hline
MetaQNN \cite{Baker2017} & 72.9 & 11.2M & 9 & 100 \\
\hline
Large Scale Evolution \cite{Real2017} & 77.0 & 40.4M & $ \ge 13$ & $ \ge 2600 $ \\
\hline
SMASH \cite{Brock2017} & 79.4 & 16M & 211 & 1.5 \\
\hline
Hill Climbing \cite{Elsken2018b} & 76.6 & 22.3M & 30 & 1 \\
\hline
NSGA-NET-128 \cite{Lu2018} & 79.3 & 3.3M & 21 & 8 \\
\hline
NSGA-NET-256 \cite{Lu2018} & 80.2 & 11.6M & 21 & 8 \\
\hline
PNAS \cite{Liu2018} & 80.5 & 3.2M & 15 & 225 \\ 
\hline
ENAS \cite{Pham2018a} & 80.6 & 4.6M & 25 & 0.45 \\
\hline
DARTS \cite{Liu2018a} & 82.5 & 3.3M & -- & 4 \\
\hline
NAONet \cite{Luo2018a} & \bfseries 84.3 & 10.8M & 30 & 200 \\
\hline
\bfseries ASED (best) & 77.3 & 16.9M & 12 & 20\\
\hline
\end{tabular}
\vspace{1pt}
\end{table*}

Modified variants of ASED prove superior to the baseline version, with the notable exception of ProbCap, which appears to stagnate at an earlier point of the search, leaving it with the smallest depth. The shortcut-based variants overall outperform other solutions for lower channel counts, but the inversion-based variants are the most efficient for larger models (in terms of both accuracy and the number of parameters). This result shows that perturbing the prototype, even very aggressively, can lead to discovering better solutions with the same or smaller depth. The shortcuts are essential in reaching deeper models without premature convergence; such models are immediately more powerful and offer short-term performance advantages, but their larger sizes imply higher memory and computation requirements. Both types of the proposed techniques also prove essential in increasing the advantage over the random search, which is otherwise quite competitive with the baseline variant of ASED.

The experiments also demonstrate the impact of brief training regime on the search process. Naturally, training a given model for less epochs usually results in lower performance; however, the algorithm design assumes that the relative ranking of candidates is still accurate, i.e., that the better structures outperform their inferiors. However, in practice this condition does not strictly hold, as can be seen by visually comparing the data from \mbox{Fig.~\ref{fig_acc}} and \mbox{Table~\ref{table_own_comparison}}. The candidate ranking is partially distorted: while overall the prototype performance increases with time, validation results can be confusing for the search. For example, the baseline ASED variant has the highest validation performance with brief training regime, but ultimately proves to be the weakest. This leaves the possibility that some promising (or even superior) models can be mistakenly discarded during the search, which would be detrimental to the final result. While this problem could be reduced by extending the brief training regime, that would also have a dramatic effect on overall computational costs; alternative solutions should be investigated. One option is low-fidelity training with a different constraint: instead of limiting the epoch number one can limit the training data size or the number of channels.  More sophisticated solutions include dynamic allocation of the training budget to prioritize more promising architectures (see e.g. \cite{Li2018b}) or predicting the candidate performance from its properties without training it (similar to \cite{Liu2018}). All of these options also have an additional benefit in speeding up the evaluation process, which is the slowest step in the ASED algorithm.

Another important observation can be made about the practical model depth achieved by different algorithm variants. If the evolution of the prototype reaches a point where the newly added layer assigns the largest probability to an identity operation, adding further layers is unlikely to introduce structural novelty, as they would also tend to converge to identities. The search essentially stops at that point, as the depth does not increase further. All of the algorithm variants that do not use shortcuts demonstrate this behaviour, although the inversion variants are capable of avoiding it to an extent by resetting the probability of identity to a low value for previously discovered layers. This effect can once again be traced back to the brief training regime and the performance estimates it produces. Fig.~\ref{fig_acc} shows how narrow the range of the estimated accuracy is. The short training schedule biases the search towards architectures that show the fastest improvement in early epochs. As a result, ASED exhibits \emph{low complexity bias}. This can be advantageous in specific cases, but it makes the architectures of higher complexity, which are naturally slower to train, much less competitive and less likely to be retained. Shortcut-based ASED circumvents this problem by making the deeper models easier to train and, therefore, more competitive. The search can thus explore more complex solutions without the induced "limit" to the layer count. However, shortcut-based ASED is still to some extent affected by the bias towards simpler architectures: Table~\ref{table_own_comparison} shows that the superiority of the models with shortcuts is limited to the smaller channel counts, which are closer to the low-fidelity evaluation setting. While the low complexity bias limits the exploration of high-dimensional structures, it can be turned into an advantage in the appropriate problem setting, e.g., if the goal is to find specifically the simplest, fastest to train models. 

The best discovered architectures (see Fig.~\ref{fig_best_nets}) can be compared with common handcrafted designs, as well as with networks that are representable by cell-based search spaces. The solutions found by ASED have pooling layers distributed roughly regularly along the network depth. This draws parallels to other common designs, where pooling (more generally, downsampling) typically separates different network submodules/blocks. However, it is important to note that the architectures discovered by ASED overall do not contain repeating sequences of operations and, therefore, cannot be represented as sequences of identical cells. These final solutions lie outside of the search space of most other architecture search methods, which demonstrates the value of generality of the proposed representation. Another notable difference between the proposed and existing solutions is the tendency for the convolution kernel size to increase along the network depth. Handcrafted designs most commonly feature larger convolutions in the earlier stages, or simply use the same kernel size for all the layers. In the ASED-produced solutions, on the other hand, 7x7 convolutions and dilated 5x5 convolutions (with an effective receptive field of 9x9) are dominant in the later layers, which is most likely the result of the low complexity bias. 

Table~\ref{table_cifar100_global} presents the comparison between the ASED algorithm (the full inversion variant, due to the highest overall performance) and a variety of existing solutions with reported results on CIFAR-100. First four rows of the table are occupied by the handcrafted architectures, while the rest are the results of architecture search. The classification accuracy and other metrics are reproduced from the corresponding papers (an empty cell indicates that the value was not originally reported). PNAS, ENAS, DARTS, and NAONet are some of the most popular and high-performing methods, but their original papers did not report results on CIFAR-100, therefore we replicate the evaluations done by other authors: we borrow DARTS results from \cite{Yu2020}, while the rest are from \cite{Luo2018a}. PNAS, ENAS, DARTS, and NAONet all utilize cell-based search spaces, while the rest of the algorithms do not. It is important to note that the results from the table were acquired under a variety of non-matching environments and settings (note that this refers not to the search methods themselves, but only to the training of the final discovered architectures). These settings may include different learning algorithms, data preprocessing, regularization, early stopping, architecture transfer (commonly from CIFAR-10 dataset), and others. Our own reproduction of the results of other algorithms under a unified setting is not possible due to the time required, as well as the lack of public original codes. We choose the follow the basic standard procedure for CIFAR-100 to establish the clear baseline, but this incurs an inherent disadvantage in the numbers against more sophisticated training techniques used by some of the competing methods. It is therefore expected that the performance of ASED can be further improved just by extra tuning of the training setup for the output network, which we leave for the future work Increasing the maximal search depth is also promising and can be achieved, for instance, by combining the shortcuts with the inversion technique. While this would lead to further increase in computation, it is worth emphasizing that ASED scales almost linearly in the parallel setting (due to the candidate models being sampled and processed independently from each other).

While ASED does not reach the state of the art in accuracy, it achieves a competitive performance overall, especially compared with the non-cell-based methods. This is noteworthy, given that the search space of ASED is relatively shallow, constantly growing in dimensionality and lacking the inherent repeating structure that greatly benefits the cell-based competitors. As previously mentioned, ASED also does not take advantage of the weight sharing or specifically tuned training procedure for its output. These facts, combined with the simple, parallelizable and extensible nature of ASED, make it a promising research direction. Given the multitude of ways to expand on the baseline solution, whether by modifying the representation or the search procedure, the proposed method is capable of discovering novel irregular architectures. The main limitation of ASED currently lies in scalability: further increasing the depth limit would result in longer candidate training, greatly increasing the overall runtime. Tackling this problem should be the priority in any future work.

\section{Conclusion}
The automated neural architecture design is growing in importance as the application-driven demand outpaces the available expertise and resources. While there are many architecture search algorithms being proposed, the compromises they have to make for the sake of performance are often limiting the range of models that can be discovered. In this paper, we proposed a novel probabilistic representation of the deep network structure and defined an architecture search algorithm, named ASED, that does not restrict the networks to repeating sequences of blocks and iteratively increases the search depth. This allows it to cover the regions of the design space which most existing approaches cannot reach, and the experiments indeed demonstrate that ASED can discover competitive non-regular architectures. However, in its current formulation ASED has difficulties when reaching larger depths, which limits its ability to construct models on the same scale as other methods. Still, the design of ASED is intuitive and extensible, which leaves many promising options for the future work and improvement. Specifically, the representation can be trivially extended by adding new layer operations, hyperparameter encodings and/or structural patterns. Computationally ASED benefits from the ease of parallel implementation, due to the independent processing of the candidate networks. The underlying search mechanism can incorporate many of the developments in the area of EDAs. Low complexity bias of ASED warrants investigation as a potentially beneficial effect, e.g., when optimizing under heavy computational constraints for execution. Finally, ASED also offers opportunities for studying the architecture search dynamics, as its internal state (prototype) is easily interpretable at any point of the search.

\ifCLASSOPTIONcaptionsoff
  \newpage
\fi

\bibliographystyle{IEEEtran}
\bibliography{references,control}

\begin{IEEEbiography}[{\includegraphics[width=1in,height=1.25in,clip,keepaspectratio]{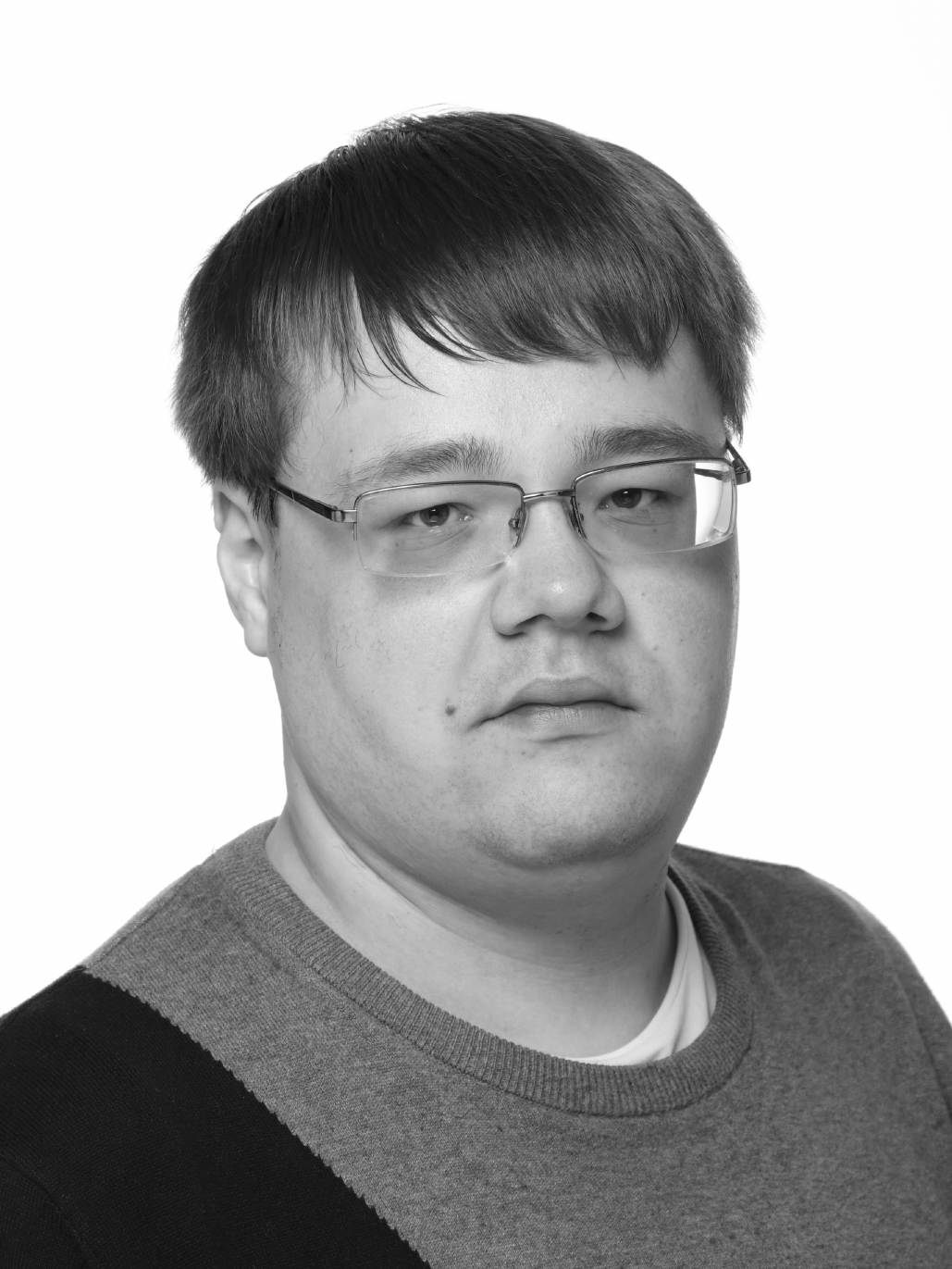}}]{Anton Muravev}
received his B.Sc. and M.Sc. degrees in computer science from the Tomsk Polytechnic University, Tomsk, Russia, followed by the M.Sc. degree in information technology from the Tampere University of Technology in 2013, 2015 and 2016 respectively. He is currently working towards the Ph.D. degree at the Tampere University, Tampere, Finland. His research interests include deep learning, pattern recognition and evolutionary computation. 
\end{IEEEbiography}
\begin{IEEEbiography}[{\includegraphics[width=1in,height=1.25in,clip,keepaspectratio]{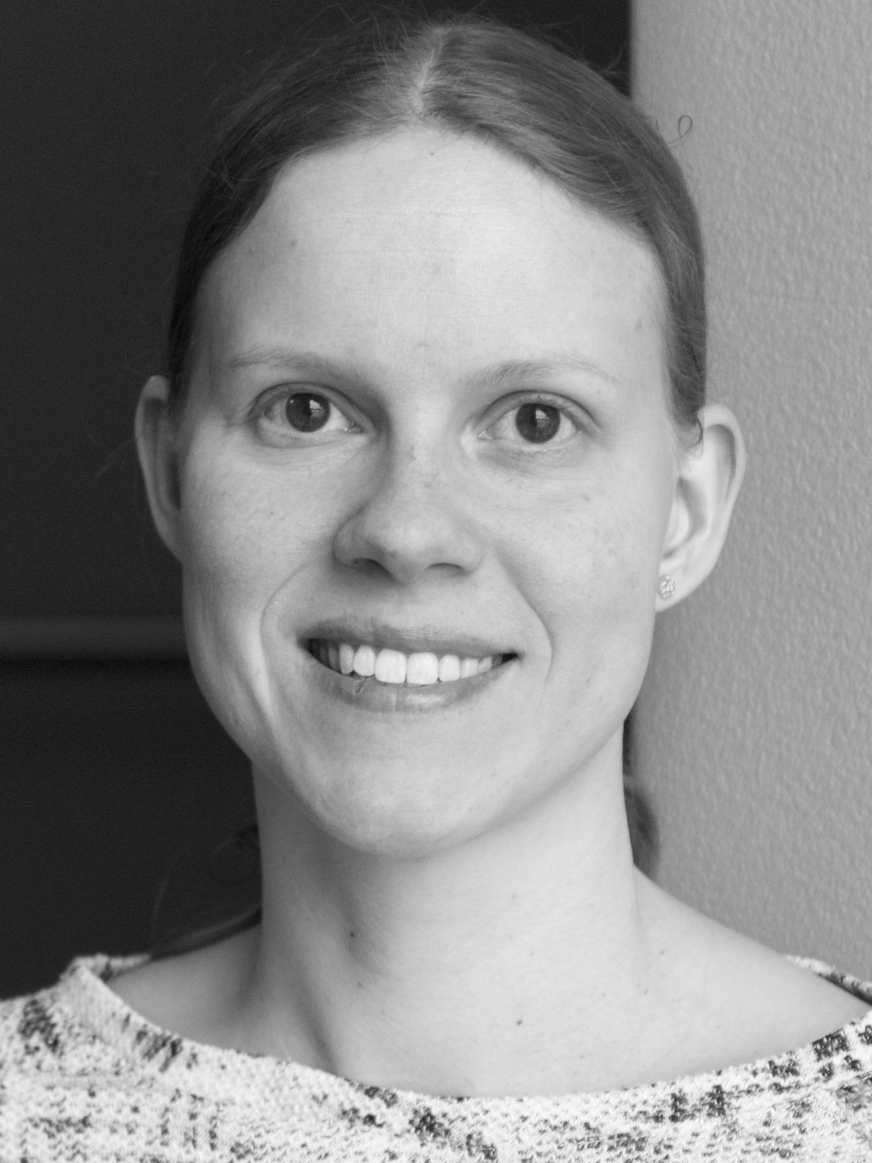}}]{Jenni Raitoharju} received her Ph.D. degree at Tampere University of Technology, Finland in 2017. She currently works as a Senior Research Scientist at the Finnish Environment Institute, Jyväskylä. She has co-authored 23 international journal papers and 32 papers in international conferences. Her research interests include machine learning and pattern recognition methods along with applications in biomonitoring and autonomous systems. She leads two research projects funded by Academy of Finland focusing on automatic taxa identification. She is the chair of Young Academy Finland 2019-2021.
\end{IEEEbiography}
\begin{IEEEbiography}[{\includegraphics[width=1in,height=1.25in,clip,keepaspectratio]{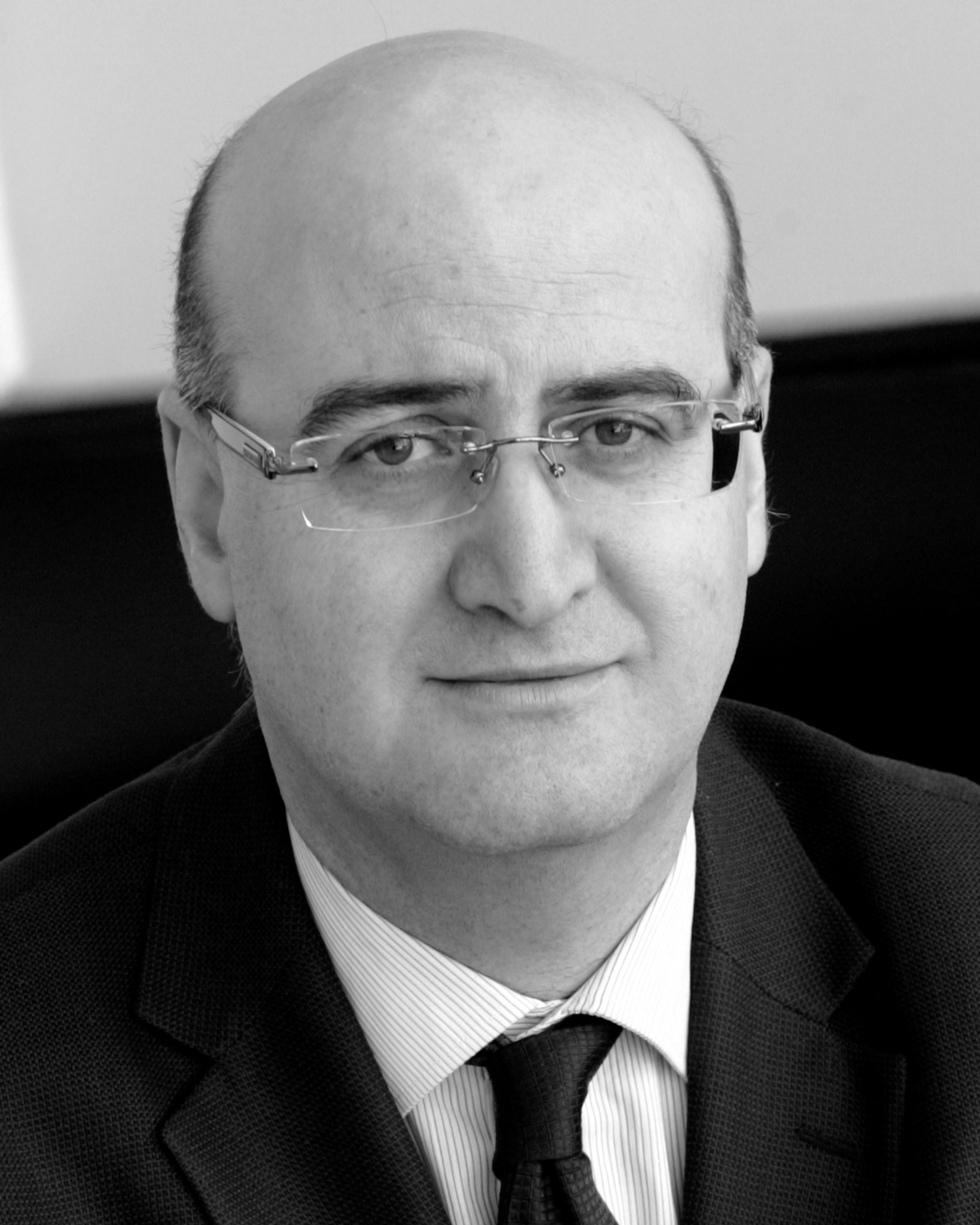}}]{Moncef Gabbouj}
received his MS and PhD degrees in electrical engineering from Purdue University, in 1986 and 1989, respectively. Dr. Gabbouj is a Professor of Signal Processing at the Department of Computing Sciences, Tampere University, Tampere, Finland. He was Academy of Finland Professor during 2011-2015. His research interests include Big Data analytics, multimedia content-based analysis, indexing and retrieval, artificial intelligence, machine learning, pattern recognition, nonlinear signal and image processing and analysis, voice conversion, and video processing and coding. Dr. Gabbouj is a Fellow of the IEEE and member of the Academia Europaea and the Finnish Academy of Science and Letters. He served as associate editor and guest editor of many IEEE, and international journals.
\end{IEEEbiography}
\EOD
\end{document}